\RequirePackage{fix-cm}
\documentclass[twocolumn]{svjour3}          
\smartqed  
\usepackage{graphicx}
\usepackage{cite}
\usepackage{amsmath,amssymb,amsfonts}
\usepackage{algorithmic}
\usepackage{textcomp}
\usepackage{subfigure} 
\newcommand{\etal}{\textit{et al.}\@ }
\newcommand{\ie}{\textit{i.e.}\@ }
\usepackage{multirow}
\usepackage{wrapfig}
\usepackage{tabularx,booktabs}
\newcolumntype{Y}{>{\centering\arraybackslash}X}

\begin{document}

\title{Effective Shortcut Technique for GAN}


\author{Seung Park \and Cheol-Hwan Yoo \and Yong-Goo Shin }

\institute{S. Park \at
              Biomedical Engineering, Chungbuk National University Hospital, 776, Seowon-gu, Cheongju-si, Chungcheongbuk-do, Rep. of Korea\\
              \email{spark.cbnuh@gmail.com}
           \and
           C.-H. Yoo \at
              Artificial Intelligence Research Laboratory, Electronics and Telecommunications Research Institute, Yuseong-gu, Daejeon, 34129, Rep. of Korea \\
              \email{ch.yoo@etri.re.kr}
           \and
           Y.-G. Shin  \at
              Division of Smart Interdisciplinary Engineering, Hannam University, Daedeok-Gu, Daejeon, 34430, Rep. of Korea\\
              \email{ygshin@hnu.kr (corresponding author)}
}

\date{Received: date / Accepted: date}

\maketitle
\begin{abstract}
In recent years, generative adversarial network (GAN)-based image generation techniques design their generators by stacking up multiple residual blocks. The residual block generally contains a shortcut, \ie skip connection, which effectively supports information propagation in the network. In this paper, we propose a novel shortcut method, called the gated shortcut, which not only embraces the strength point of the residual block but also further boosts the GAN performance. More specifically, based on the gating mechanism, the proposed method leads the residual block to keep (or remove) information that is relevant (or irrelevant) to the image being generated. To demonstrate that the proposed method brings significant improvements in the GAN performance, this paper provides extensive experimental results on the various standard datasets such as CIFAR-10, CIFAR-100, LSUN, and tiny-ImageNet. Quantitative evaluations show that the gated shortcut achieves the impressive GAN performance in terms of Frechet inception distance (FID) and Inception score (IS). For instance, the proposed method improves the FID and IS scores on the tiny-ImageNet dataset from 35.13 to 27.90 and 20.23 to 23.42, respectively.  

\keywords{Generative adversarial networks \and image generation \and residual block \and gated shortcut}
\end{abstract}

\section{Introduction}
Generative adversarial networks (GAN)~\cite{goodfellow2014generative} is an algorithmic framework that shows impressive performance in diverse applications including image-to-image translation~\cite{isola2017image, choi2018stargan, zhu2017unpaired}, text-to-image translation~\cite{reed2016generative, hong2018inferring}, and image inpainting~\cite{yu2018free, sagong2019pepsi, shin2020pepsi++}. Generally, the GAN consists of two different networks, called generator and discriminator, which are trained with opposite goals. The generator is trained to fool the discriminator by synthesizing the sample following the real data distribution, whereas the goal of the discriminator is to classify the real and generated samples. Since the networks compete with each other, GAN is more difficult to train stably than the supervised learning-based convolutional neural networks (CNNs)~\cite{zhang2019consistency}. To alleviate this problem, some papers~\cite{zhang2019self, karras2017progressive, zhang2018stackgan++} proposed novel network architectures for the generator and discriminator. Although these approaches could generate high-resolution images on challenging datasets such as ImageNet~\cite{deng2009imagenet}, they still cannot fully resolve the instability problem during the training procedure~\cite{park2021generative}.

Recent studies pointed out that the major reason for the instability problem is the sharp gradient space of the discriminator~\cite{wu2021gradient}. To mitigate this issue, some papers~\cite{miyato2018spectral, gulrajani2017improved, kodali2017convergence, zhang2019consistency, kurach2019large} proposed normalization or regularization methods that restrain the sharp gradient space. In particular, as a normalization method, spectral normalization (SN)~\cite{miyato2018spectral} is the most popular method that constrains the Lipschitz constant of the discriminator by dividing weight matrices with an approximation of the largest singular value. As the regularization technique, the conventional works~\cite{gulrajani2017improved, wu2019generalization, wei2018improving, kodali2017convergence, roth2017stabilizing} usually added the regularization term into the adversarial loss function. Among the diverse regularization methods, gradient penalty-based regularization methods~\cite{gulrajani2017improved, wu2019generalization, wei2018improving}, which constrains the magnitude of gradient, have been widely used. Since these normalization and regularization techniques can lead the GAN training stably, the recent works generally employ these techniques when training their networks. 

Some papers introduced novel architectural units utilized for improving the GAN performance. Miyato~\etal~\cite{miyato2018cgans} proposed a conditional projection technique that projects the feature vector onto the conditional weight vector to provide conditional information to the discriminator. Since this approach is effective to boost the conditional GAN (cGAN) performance, it is widely used in recent studies. Zhang~\etal~\cite{zhang2019self} applied a self-attention mechanism, which guides the generator and discriminator where to attend, to produce high-quality images, whereas Yeo~\etal~\cite{yeo2021simple} proposed a cascading rejection module that classifies real and generated samples using dynamic features in an iterative manner at the last layer of the discriminator. On the other hand, some studies proposed new forms of convolution layer which is effective to improve the GAN performance. Sagong~\etal~\cite{sagong2019cgans} introduced a conditional convolution (cConv) for the generator, which provides the conditional information to the convolution operation via scaling and shifting operations. cConv is able to improve the cGAN performance well but it has a major drawback: cConv only can be utilized for cGAN scheme since it is developed for replacing the conditional batch normalization. Park~\etal~\cite{park2021generative} proposed a perturbed convolution (PConv) that not only prevents the discriminator from falling into the overfitting problem but also enhances the GAN performance. However, since PConv is designed for the discriminator, it is difficult to employ PConv to the generator. Recently, Park~\etal~\cite{park2021generative_conv} introduced a generative convolution that modulates the convolution weight of the generator following the given latent vector. 

Most state-of-the-art studies~\cite{miyato2018spectral, miyato2018cgans, brock2018large, zhang2019self, park2021generative_conv, park2021generative, yeo2021simple, sagong2019cgans} built their generators by stacking up multiple residual blocks~\cite{he2016deep, he2016identity} consisting of convolution layer, batch normalization (BN)~\cite{ioffe2015batch}, and activation function. More specifically, these studies investigated the training strategies or architectural units without redesigning or modifying the residual block. Recently, some papers tackled these approaches and introduced novel residual block for improving the GAN performance. Wang~\etal~\cite{wang2021up} proposed new from of the residual block, called UpRes, which modifies the block architecture as well as the activation function. However, as pointed out in~\cite{park2021GRB}, the performance improvement of UpRes is caused by the activation function, not caused by its architecture. Park~\etal~\cite{park2021GRB} introduced a generative residual block (GRB) which contains a side-bottleneck path designed for effectively emphasizing the informative feature while suppressing the less useful one. Although the GRB shows superior performance to the residual block, designing a novel residual block still remains an open question. 

Among the various parts of the residual block, we mainly focus on a shortcut, \ie skip connection, which is designed for aiding the information propagation in the network. In~\cite{he2016identity}, they proved that an identity shortcut, which adds the input feature directly to the output one, is the most effective way for the image classification task. Indeed, the network for the image classification task is trained to cluster high-level features that represents the characteristics of the image. However, the goal of the generator is a little different. To generate various images, the generator should learn how to well disperse the high-level features (latent features) to the diverse low-level ones (image). Since the goals of the image classification network and generator are opposed to each other, we suspect that the identity shortcut may not the best choice for the generator. To validate our hypothesis, this paper proposes a novel skip connection method, called the gated shortcut, which embraces the strength point of the residual block as well as further enhances the GAN performance. By using the gating mechanism, the proposed method supports the residual block to keep (or remove) information that is relevant (or irrelevant) to the image being generated. In other words, unlike with the identity shortcut, the proposed method propagates the selected information in the input feature to the output feature. To prove that the proposed method brings significant improvements in the GAN performance, we performed extensive experimental results on various datasets including CIFAR-10~\cite{krizhevsky2009learning}, CIFAR-100~\cite{krizhevsky2009learning}, LSUN~\cite{yu15lsun}, and tiny-ImageNet~\cite{deng2009imagenet, yao2015tiny}. Quantitative evaluations show that the proposed method accomplishes the remarkable GAN and conditional GAN (cGAN) performance in terms of Frechet inception distance (FID) and Inception score (IS). 

In summary, key contributions of our paper are as follows: First, we propose a novel shortcut method that supports the residual block to keep (or remove) information that is relevant (or irrelevant) to the image being generated. Second, with slight additional network parameters, the proposed method significantly improves the GAN and cGAN performances on various datasets in terms of IS and FID. For example, the proposed method boosts FID and IS scores on the tiny-ImageNet dataset from 35.13 to 27.90 and 20.23 to 23.42, respectively.  

\section{Background}
\label{sec2}
\subsection{Generative Adversarial Networks}
\label{sec2.1}
In general, GAN~\cite{goodfellow2014generative} consists of two different networks, \ie generator $G$ and discriminator $D$. $G$ is optimized to produce visually plausible images, whereas $D$ is trained to classify the generated images from real ones. Since the goals of those network are opposed to each other, we call this training procedure adversarial learning. The loss function for adversarial learning is defined as follows:
\begin{eqnarray}
\label{eq1:ganD}
       \lefteqn{L_D = -E_{x\sim P_\textrm{data}(x)}[\log D(x)]}\nonumber\\
    & {\qquad \qquad \qquad} - E_{z\sim P_{z(z)}}[\log(1-D(G(z)))],
\end{eqnarray}
\begin{eqnarray}
\label{eq1:ganG}
    L_G = -E_{z\sim P_{z(z)}}[\log(D(G(z)))],
\end{eqnarray}
where $L_D$ and $L_G$ indicate the loss function for the discriminator and generator, respectively, whereas $z$ and $x$ are the random vector sampled from the Gaussian normal distribution $P_z(z)$ and a image sampled from the real image distribution $P_{data}(x)$, respectively. To make adversarial learning stably, some papers~\cite{mao2017least, arjovsky2017wasserstein, lim2017geometric, gulrajani2017improved} attempted to reformulate the loss function. Among the various loss functions, a hinge-adversarial loss~\cite{lim2017geometric} is widely used in practice~\cite{yeo2021simple, miyato2018cgans, miyato2018spectral, brock2018large, park2021generative, sagong2019pepsi, shin2020pepsi++, chen2019self, park2021novel}. On the other hand, cGAN that produces the class-conditional images has also been actively studied~\cite{mirza2014conditional, odena2017conditional, miyato2018cgans}. cGAN often utilizes additional conditional information such as class labels or text conditions, in order to produce the condition-specific image. By minimizing the loss functions for cGAN~\cite{mirza2014conditional, miyato2018cgans}, the user can select the image to be generated. The reader is encouraged to review the adversarial learning of GAN and cGAN for more details.

\begin{figure}
\centering
\includegraphics[width=0.9\linewidth]{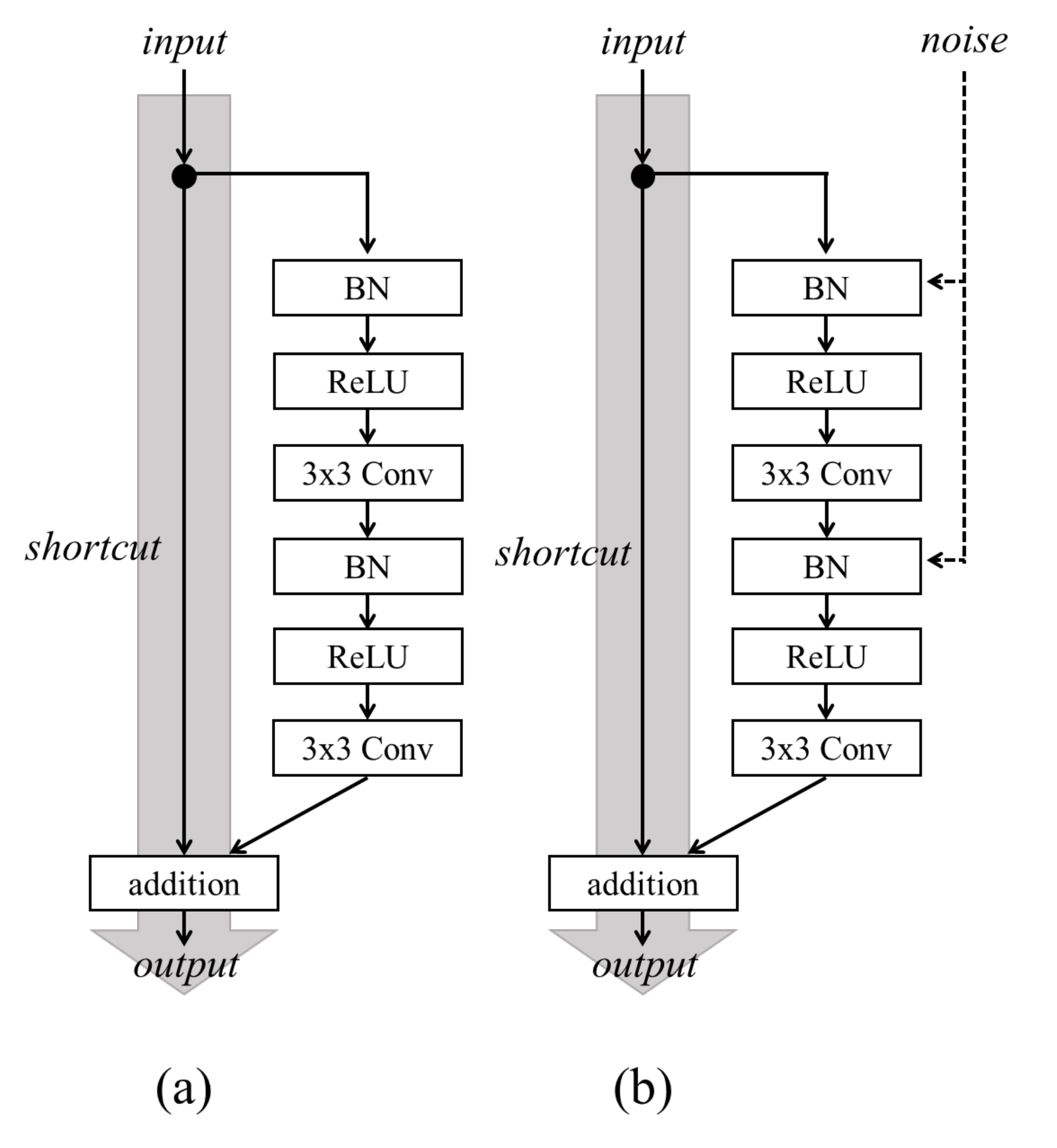}
\caption{Residual blocks in the previous methods. (a) Residual block in SNGAN~\cite{miyato2018spectral}, (b) Residual block in BigGAN~\cite{brock2018large}.}
\label{fig:fig1}
\vspace{0cm}
\end{figure}

\subsection{Residual block for GAN}
\label{sec2.2}
The residual block~\cite{he2016deep} was first developed for mitigating the gradient vanishing problem that occurs in deep CNNs. Since the residual block exhibits fine performance in various applications and is easy to implement, recent GAN studies~\cite{miyato2018spectral, miyato2018cgans, brock2018large, zhang2019self, park2021generative_conv, park2021generative, yeo2021simple, sagong2019cgans} naturally build their generator using the residual block. Specifically, in~\cite{miyato2018spectral}, they introduced impressive GAN framework, called SNGAN, which builds the generator by stacking the standard residual block shown in Fig.~\ref{fig:fig1}(a). To further improve the GAN performance, as shown in Fig.~\ref{fig:fig1}(b), Brock~\etal~\cite{brock2018large} used a modified residual block in which the scaling and shifting parameters of the batch normalization are inferred from the noise vector. Since these residual blocks show fine performance in the GAN-based image generation task, recent studies used these residual blocks to build their generator.

\begin{figure*}
\centering
\includegraphics[width=0.85\linewidth]{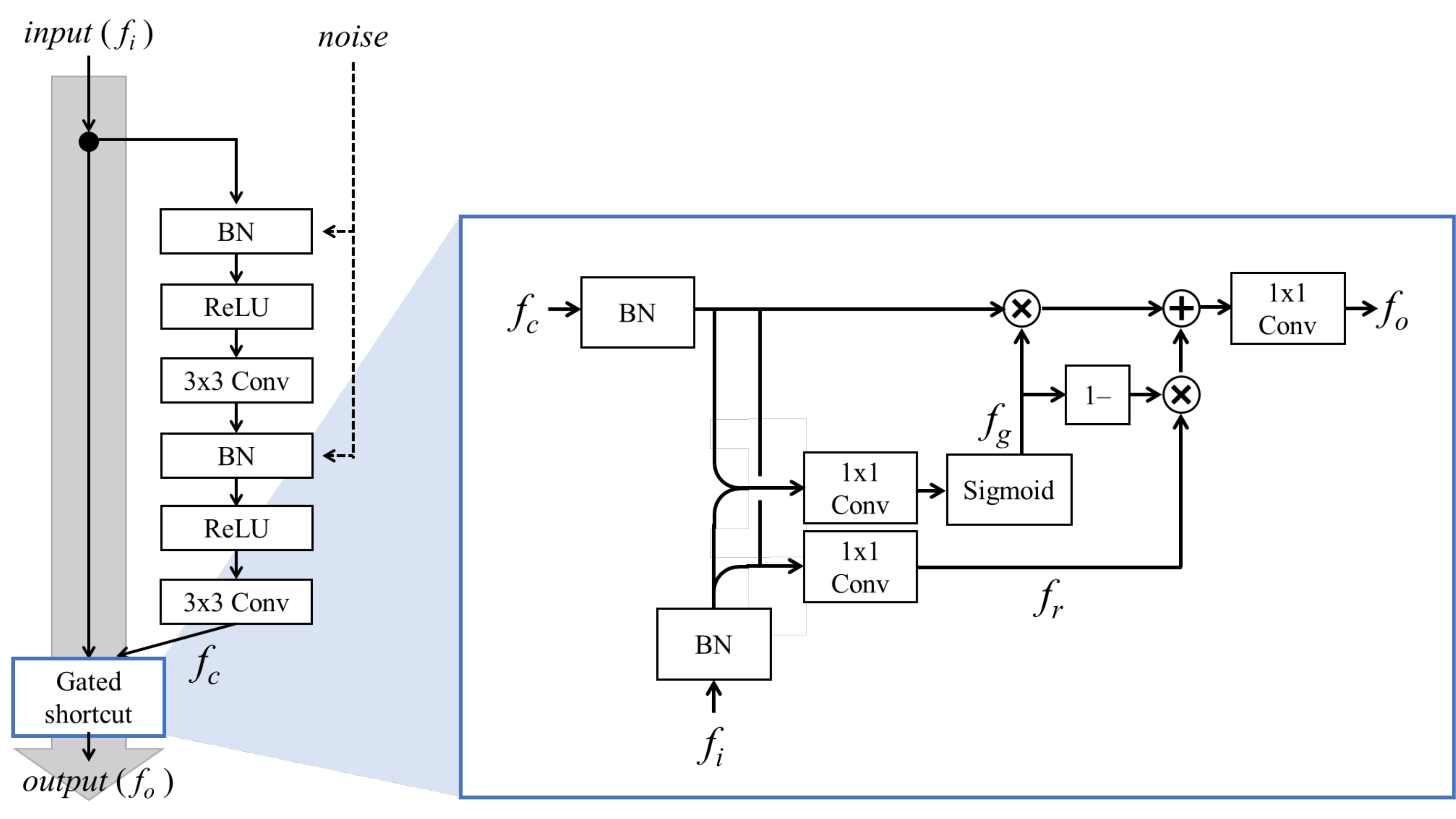}
\caption{Overall structure of the proposed residual block. Like the residual block in BigGAN~\cite{brock2018large}, we produces the affine transformation parameters of BN using the noise vector. Instead of using the identity shortcut, the proposed method employs the proposed gated shortcut.}
\label{fig:fig2}
\vspace{-0cm}
\end{figure*}

The common characteristic of these blocks is that they used the identity shortcut which adds the input feature to the output one without any modifications. The identity shortcut is widely employed in various CNN-based applications since He~\etal~\cite{he2016identity} proved its effectiveness by conducting various ablation studies. Indeed, the effectiveness of the identity shortcut has been only proved for the image classification task, not for GAN-based image generation technique; the effectiveness of the identity shortcut for the generator did not demonstrate in that paper~\cite{he2016identity}. The goal of generator is to produce the image using the high-level feature in the latent space. That means, to generate diverse images, each residual block should learn how to well project the input high-level feature onto the output low-level feature. Here, we suspect that there is an issue when using the identity shortcut. When employing the identity shortcut, the problem could occur since all information in the high-level feature is added to the projected low-level feature; there could be more effective shortcut technique that alleviates this concern. To prove this hypothesis, this paper introduces a novel shortcut method, called gated shortcut, which significantly improves the GAN performance.

\section{Proposed Method}
\label{sec3}
\subsection{Gated Shortcut}
\label{subsec3.1}
Fig.~\ref{fig:fig2} shows the overall structure of the proposed residual block. In the proposed method, like the residual block in BigGAN~\cite{brock2018large}, we infer the affine transformation parameters of BN using the noise vector. However, instead of using the identity shortcut, the proposed method employs the proposed gated shortcut that supports the residual block to keep (or erase) information in the input feature, which is relevant (or irrelevant) to the output feature. Specifically, in the proposed residual block, we first produce the feature $f_c\in\mathbb{R}^{h\times w \times c_c}$ using multiple convolutional layers, which represents lower-level feature than the input feature $f_i\in\mathbb{R}^{h\times w \times c_i}$. Here, $h$ and $w$ mean the height and width of the feature map, respectively, whereas $c_\textrm{x}$ indicates the channel numbers of $f_\textrm{x}$. Then, to perform the residual learning, we combine $f_i$ and $f_c$ using the proposed gated shortcut. 

To effectively handle that which information in $f_i$ will be kept or removed, we design the gated shortcut using the gating mechanism~\cite{hochreiter1997long}. In the gated shortcut, since the feature scales of $f_c$ and $f_i$ are different, we first balance the scales using BN~\cite{ioffe2015batch}. Then, to conduct the gating mechanism, we first produce the gate $f_g \in\mathbb{R}^{h\times w \times c_g}$ as follows:
\begin{equation}
    f_g = \sigma \big[\textrm{W}_g*(f_c \odot f_i)\big],
\label{eq3}
\end{equation}
where $\textrm{W}_g$ indicates a trainable weight in the convolution layer, whereas $*$ and $\odot$ mean the convolution and channel concatenation operations, respectively. $\sigma$ is a sigmoid function that constrains output in the range [0, 1] to control how much to keep or forget the information. In addition, we compute the refinement feature $f_r \in\mathbb{R}^{h\times w \times c_r}$, which will be newly added to $f_c$. That means we attempt to only take the meaningful information in $f_i$. The $f_r$ is computed as
\begin{equation}
    f_r = \textrm{W}_r*(f_c \odot f_i),
\label{eq4}
\end{equation}
where $\textrm{W}_r$ is a trainable weight in the convolution layer. Consequently, the output feature $f_o \in \mathbb{R}^{h\times w \times c_o}$ can be computed by using \ie $f_g$ and $f_r$ as follows:
\begin{equation}
    f_o = \textrm{W}_o*\big[f_g \otimes f_c + (1 - f_g) \otimes f_r\big],
\label{eq5}
\end{equation}
where $\otimes$ indicates element-wise multiplication and $\textrm{W}_o$ is a weight in the output convolution layer. The arithmetic operations in Eq.~\ref{eq5} could be elucidated as follows: the first term $f_g \otimes f_c$ means that the gate decides what information needs to memorize in the $f_c$, whereas the second term $(1 - f_g) \otimes f_r$ represents that $f_c$ is refined with selected information coming from $f_i$. That means, the network can effectively control information flow by adding only necessary information derived from $f_i$ to $f_c$. 

Here, one may anticipate that the gated shortcut is similar with the exclusive gating shortcut (EGS) or shortcut only gating (SOG) that are used for ablation studies in~\cite{he2016identity}. However, there is a major difference between the proposed method and conventional ones. For instance, the EGS computes output $f_{EGS}$ by summing the input and convolution features, \ie $f_i$ and $f_c$, after multiplying the gate $f_g$ as follows: 
\begin{equation}
    f_{EGS} = f_g \otimes f_i + (1 - f_g) \otimes f_c.
\label{eq6}
\end{equation}
In Eq.~\ref{eq6}, $f_{EGS}$ can be interpreted as the weighted summation of the $f_i$ and $f_c$, where weight value is $f_g$. Thus, if $f_g$ is 0.5, it is equal to the scaled-identity shortcut; it cannot effectively keep (or remove) the relevant (or irrelevant) information in $f_c$. In contrast, instead of directly summing $f_i$, the proposed method produces the refinement feature, \ie $f_r$. It is worth noting that this simple procedure derives significant performance improvement. The detailed explanations and ablation studies will be introduced in the Section~\ref{subsec:4.2}. 

\subsection{Implementation Details}
\label{subsec:3.2}
To show the effectiveness of the proposed method, this paper presents extensive experimental results using the various standard datasets used for the GAN study. In our experiments, we employed CIFAR-10~\cite{krizhevsky2009learning}, CIFAR-100~\cite{krizhevsky2009learning}, LSUN~\cite{yu15lsun}, and tiny-ImageNet~\cite{deng2009imagenet, yao2015tiny} which is a subset of ImageNet~\cite{deng2009imagenet}, consisting of the 200 selected classes. More specifically, among the various classes in the LSUN dataset, we only used the church and tower images. The images in the CIFAR-10 and CIFAR-100 datasets are $32\times 32$ pixels, whereas the images in LSUN and tiny-ImageNet datasets are resized to $128\times 128$ pixels. 

We utilized the hinge-adversarial loss\cite{lim2017geometric} to train GAN. In addition, since all parameters in the discriminator and generator including the gated shortcut can be differentiated, we employed the Adam optimizer~\cite{kingma2014adam} and set the user parameters, \ie $\beta _1$ and $\beta _2$, to 0 and 0.9, respectively. For training the CIFAR-10 and CIFAR-100 datasets, we set the learning rate as 0.0002 and updated the discriminator five times using different mini-batches when the generator is updated once. The training procedure was finished after the generator is updated 50k iterations. Following the previous studies~\cite{miyato2018cgans, miyato2018spectral, park2021GRB}, the batch size of the generator was set to twice larger than the discriminator; the batch size of the discriminator and generator is 64 and 128, respectively. 

On the other hand, for training the LSUN and tiny-ImageNet datasets, we employed a two-time scale update rule technique (TTUR)~\cite{heusel2017gans}, which set the learning rates of the generator and discriminator to 0.0001 and 0.0004, respectively. It is worth noting that TTUR method updates the discriminator a single time when the generator is updated once. The batch sizes of the discriminator and the generator are the same; we set batch sizes of both networks to 32. The generator was updated to 300k iterations on the LSUN dataset and 1M iterations on the tiny-ImageNet dataset. For all datasets, the learning rate was linearly decayed in the last 50,000 iterations. In addition, the spectral normalization (SN)~\cite{miyato2018spectral} was only used for the discriminator when training the CIFAR-10 and CIFAR-100 datasets, whereas the SN was applied to the generator as well as discriminator when training the LSUN and tiny-ImageNet datasets. 

\begin{table}[t]
\caption{Network architecture of the generator for each image resolution. The input latent vector is sample from $N(0, I)$.}
\begin{center}
\begin{tabular}{ | c | c | }
\hline
$32\times32$ image& $128\times128$ image \\
\hline
FC, $4 \times 4 \times 256$ & FC, $4 \times 4 \times 512$ \\
ResBlock, up, 256 & ResBlock, up, 512\\
ResBlock, up, 256 & ResBlock, up, 512\\
ResBlock, up, 256 & ResBlock, up, 256\\
BN, ReLU & ResBlock, up, 128 \\
$3\times3$ conv, Tanh & ResBlock, up, 64\\
& BN, ReLU\\
& $3\times3$ conv, Tanh\\

\hline
\end{tabular}
\end{center}
\label{table:table_G}
\end{table}

\begin{table}[t]
\caption{Network architecture of the discriminator for each image resolution.}
\begin{center}
\begin{tabular}{| c | c | }
\hline
$32\times32$ image & $128\times128$ image\\
\hline
ResBlock, down, 128 & ResBlock, down, 64\\
ResBlock, down, 128 & ResBlock, down, 128\\
ResBlock, 128 & ResBlock, down, 256\\
ResBlock, 128 & ResBlock, down, 512\\
ReLU & ResBlock, down, 512\\
Global Sum & ResBlock, 512\\
Dense, 1 & ReLU\\
 & Global Sum\\
 & Dense, 1\\

\hline
\end{tabular}
\end{center}
\label{table:table_D}
\end{table}

\begin{table*}[t]
\caption{Comparison of the GAN performance between the proposed method with conventional ones on the CIFAR-10 and CIFAR-100 datasets in terms of IS and FID.}
\begin{center}
\begin{tabularx}{0.95\textwidth}{ |c| *{10}{Y|} }

\hline
\multicolumn{2}{|c|}{Network} & \multicolumn{2}{c|}{SNGAN~\cite{miyato2018cgans}} & \multicolumn{2}{c|}{BigGAN~\cite{brock2018large}} & \multicolumn{2}{c|}{GRB-GAN~\cite{park2021GRB}} & \multicolumn{2}{c|}{Proposed}\\
\hline
Dataset & & IS$\uparrow$ & FID$\downarrow$ & IS$\uparrow$ & FID$\downarrow$ & IS$\uparrow$ & FID$\downarrow$ & IS$\uparrow$ & FID$\downarrow$ \\
\hline
\multirow{5}*{CIFAR-10} & trial 1 & 7.79 & 13.81 & 7.70 & 13.98 & 7.94 & 12.22 & 7.93 & 10.94 \\
& trial 2 & 7.76 & 13.29 & 7.80 & 13.69 & 7.93 & 12.45 & 7.95 & 11.14 \\
& trial 3 & 7.76 & 13.29 & 7.87 & 13.09 & 8.03 & 12.50 & 8.00 & 10.88 \\
\cline{2-10}
& Mean & 7.77 & 13.46 & 7.79 & 13.58 & 7.97 & 12.39 & 7.96 & 10.98 \\
\cline{2-10}
& Std & 0.02 & 0.30 & 0.08 & 0.45 & 0.05 & 0.15 & 0.04 & 0.14 \\
\hline
\multirow{5}*{CIFAR-100} & trial 1 & 8.08 & 17.76 & 7.99 & 17.45 & 8.29 & 16.85 & 8.22 & 15.17 \\
& trial 2 & 8.08 & 17.93 & 7.94 & 17.72 & 8.30 & 16.34 & 8.31 & 15.00 \\
& trial 3 & 8.05 & 17.55 & 7.94 & 17.57 & 8.18 & 16.64 & 8.33 & 15.39 \\
\cline{2-10}
& Mean & 8.07 & 17.75 & 7.96 & 17.58 & 8.26 & 16.61 & 8.29 & 15.19 \\
\cline{2-10}
& Std & 0.02 & 0.19 & 0.03 & 0.13 & 0.06 & 0.26 & 0.06 & 0.19 \\

\hline
\end{tabularx}
\end{center}
\label{table1}
\end{table*}

\begin{table*}[t]
\caption{Comparison of the cGAN performance between the proposed method with conventional ones on the CIFAR-10 and CIFAR-100 datasets in terms of IS and FID.}
\begin{center}
\begin{tabularx}{0.95\textwidth}{ |c| *{10}{Y|} }

\hline
\multicolumn{2}{|c|}{Network} & \multicolumn{2}{c|}{SNGAN~\cite{miyato2018cgans}} & \multicolumn{2}{c|}{BigGAN~\cite{brock2018large}} & \multicolumn{2}{c|}{GRB-GAN~\cite{park2021GRB}} & \multicolumn{2}{c|}{Proposed}\\
\hline
Dataset & & IS$\uparrow$ & FID$\downarrow$ & IS$\uparrow$ & FID$\downarrow$ & IS$\uparrow$ & FID$\downarrow$ & IS$\uparrow$ & FID$\downarrow$ \\
\hline
\multirow{5}*{CIFAR-10} & trial 1 & 8.06 & 10.26 & 8.09 & 9.36 & 8.16 & 9.86 & 8.19 & 8.14 \\
& trial 2 & 8.02 & 10.37 & 7.97 & 9.62 & 8.17 & 9.06 & 8.32 & 7.85 \\
& trial 3 & 7.99 & 10.01 & 8.02 & 9.38 & 8.22 & 9.38 & 8.26 & 8.16 \\
\cline{2-10}
& Mean & 8.03 & 10.21 & 8.03 & 9.45 & 8.18 & 9.44 & 8.26 & 8.05 \\
\cline{2-10}
& Std & 0.03 & 0.18 & 0.06 & 0.15 & 0.03 & 0.41 & 0.07 & 0.14 \\
\hline
\multirow{5}*{CIFAR-100} & trial 1 & 8.62 & 14.62 & 8.88 & 13.17 & 9.14 & 12.76 & 9.36 & 10.10 \\
& trial 2 & 8.62 & 14.39 & 8.89 & 13.20 & 8.94 & 13.20 & 9.34 & 10.74 \\
& trial 3 & 8.80 & 14.85 & 8.87 & 13.01 & 9.18 & 11.50 & 9.34 & 10.52 \\
\cline{2-10}
& Mean & 8.68 & 14.62 & 8.88 & 13.13 & 9.09 & 12.49 & 9.34 & 10.45 \\
\cline{2-10}
& Std & 0.10 & 0.23 & 0.01 & 0.10 & 0.13 & 0.88 & 0.01 & 0.33 \\

\hline
\end{tabularx}
\end{center}
\label{table2}
\end{table*}

In our experiments, we built generator and discriminator architectures following a strong baselines in~\cite{miyato2018cgans, brock2018large}, called SNGAN and BigGAN. The detailed discriminator and generator architectures are described in Tables~\ref{table:table_G} and~\ref{table:table_D}. In the discriminator, the feature maps were down-sampled by using the average-pooling layer after the second convolution layer. In the generator, the up-sampling (a nearest-neighbor interpolation) operation was located before the first convolution. The more detailed residual block architectures are explained in~\cite{miyato2018cgans, brock2018large}. In the gated shortcut, we set $c_g = c_r = c_o$, and $c_o$ for each residual block is described in Table~\ref{table:table_G}. In addition, we matched the spatial sizes of $f_i$ with $f_c$ using the nearest-neighbor interpolation after BN of $f_i$. On the other hand, to train the network in the cGAN framework, we substituted BN in the generator with the conditional BN (cBN)~\cite{dumoulin2017learned, brock2018large} and added the conditional projection layer at the last of the discriminator~\cite{miyato2018cgans}. 

\section{Experimental Results}
\label{sec4}
\subsection{Evaluation Metric}
\label{subsec:4.1}
FID~\cite{heusel2017gans} and IS~\cite{salimans2016improved} are the most prevalent metrics that evaluate how well the network generates the image. In particular, FID, which measures the Wasserstein distance between the feature distributions of the real and generated images is defined as follows: 
\begin{equation}
    \textrm{F}(p,q) = \| \mu_p - \mu_q \|_2^2 + \mathrm{trace}(C_p +C_q - 2(C_p C_q)^{\frac{1}{2}}),
\end{equation}
where $ \{\mu_p,C_p \}$ and $\{\mu_q,C_q \}$ are the mean and covariance of the distributions of real and generated images, respectively. Lower FID score indicates that the generated images have better quality. On the other hand, Salimans~\etal~\cite{salimans2016improved} revealed that IS is strongly correlated with the subjective human judgment of image quality. IS is expressed as 
\begin{equation}
I = \mathrm{exp}(E[D_{KL}(p(l|X)||p(l))]),
\end{equation}
where \textit{l} is the class label predicted by the Inception model~\cite{szegedy2016rethinking} trained by using the ImageNet dataset~\cite{deng2009imagenet}, and $p(l|X)$ and $p(l)$ represent the conditional class distributions and marginal class distributions, respectively. Contrary to the FID, the generated image with high quality achieve the high IS score. In this paper, we randomly generated 50,000 samples and measured the FID and IS scores using the same number of the real images.

\begin{table*}[t]
\caption{Comparison of the GAN performance between the proposed method with conventional ones on the CIFAR-10 and CIFAR-100 datasets in terms of IS and FID.}
\begin{center}
\begin{tabularx}{0.95\textwidth}{ |c| *{10}{Y|} }

\hline
\multicolumn{2}{|c|}{Network} & \multicolumn{2}{c|}{SNGAN~\cite{miyato2018cgans}} & \multicolumn{2}{c|}{BigGAN~\cite{brock2018large}} & \multicolumn{2}{c|}{GRB-GAN~\cite{park2021GRB}} & \multicolumn{2}{c|}{Proposed}\\
\hline
\multicolumn{2}{|c|}{Parameters} & \multicolumn{2}{c|}{4.66M} & \multicolumn{2}{c|}{4.66M} & \multicolumn{2}{c|}{4.66M} & \multicolumn{2}{c|}{4.66M}\\
\hline
Dataset & & IS$\uparrow$ & FID$\downarrow$ & IS$\uparrow$ & FID$\downarrow$ & IS$\uparrow$ & FID$\downarrow$ & IS$\uparrow$ & FID$\downarrow$ \\
\hline
\multirow{5}*{CIFAR-10} & trial 1 & 7.80 & 13.59 & 7.79 & 14.17 & 7.99  & 12.08 & 7.93 & 10.94 \\
& trial 2 & 7.77 & 13.96 & 7.74 & 12.75 & 7.98 & 12.05 & 7.95 & 11.14 \\
& trial 3 & 7.84 & 12.98 & 7.93 & 12.03 & 8.03 & 11.60 & 8.00 & 10.88 \\
\cline{2-10}
& Mean & 7.81 & 13.51 & 7.82 & 12.98 & 8.00 & 11.91 & 7.96 & 10.98 \\
\cline{2-10}
& Std & 0.04 & 0.50 & 0.10 & 1.09 & 0.02 & 0.27 & 0.04 & 0.14 \\
\hline
\multirow{5}*{CIFAR-100} & trial 1 & 8.10 & 16.95 & 8.07 & 16.34 & 8.17 & 17.77 & 8.22 & 15.17 \\
& trial 2 & 8.08 & 17.45 & 8.10 & 16.78 & 8.24 & 18.17 & 8.31 & 15.00 \\
& trial 3 & 8.17 & 17.52 & 8.06 & 16.84 & 8.16 & 17.13 & 8.33 & 15.39 \\
\cline{2-10}
& Mean & 8.12 & 17.30 & 8.08 & 16.65 & 8.19 & 17.69 & 8.29 & 15.19 \\
\cline{2-10}
& Std & 0.05 & 0.31 & 0.02 & 0.27 & 0.04 & 0.52 & 0.06 & 0.19 \\

\hline
\end{tabularx}
\end{center}
\label{table3}
\end{table*}

\begin{figure*}
\centering
\includegraphics[width=0.95\linewidth]{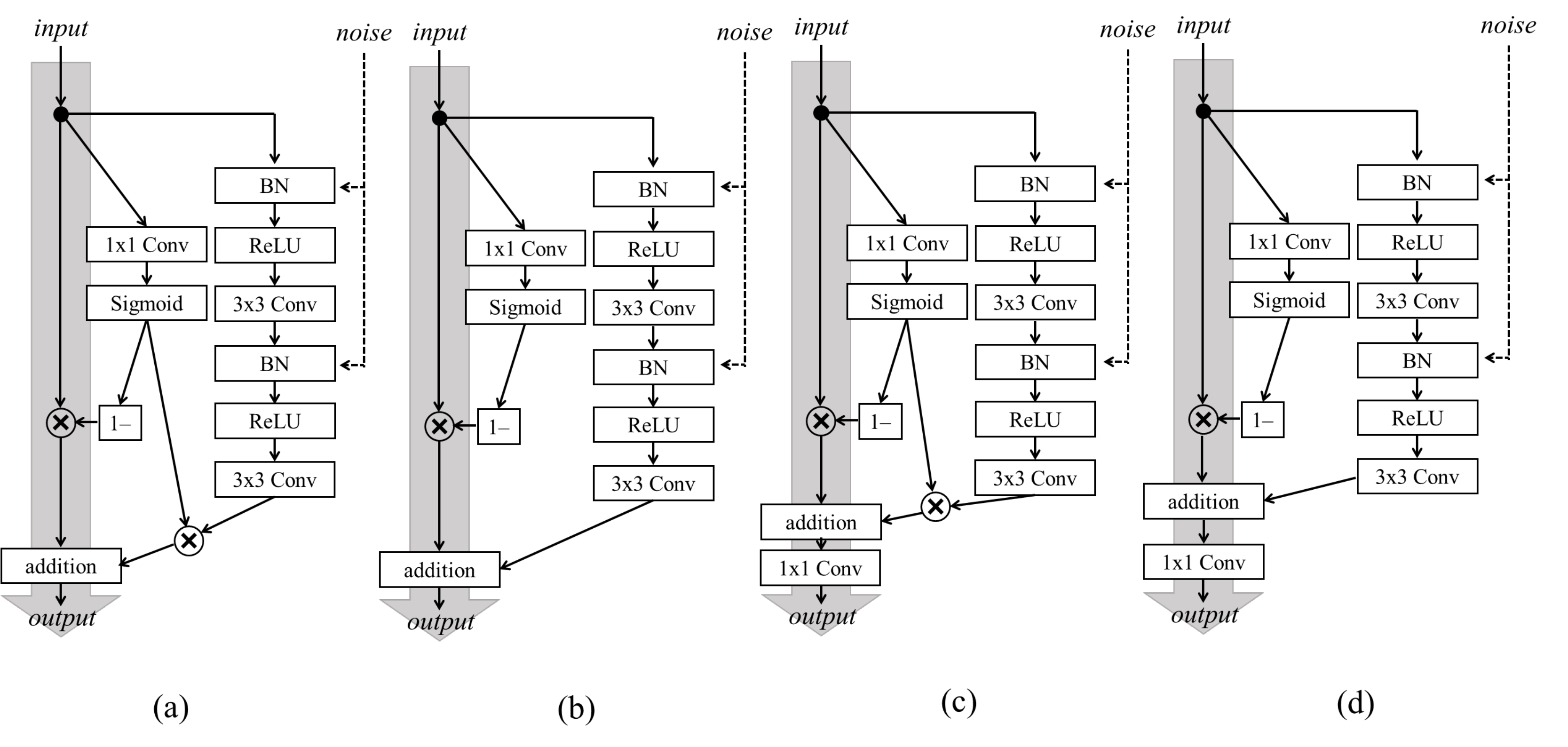}
\caption{Architectures of the residual block used for ablation studies, which are introduced in~\cite{he2016identity}. (a) Residual block with EGS, (b) Residual block with SOG, (c) Residual block with EGS and additional convolution layer, (d) Residual block with SOG and additional convolution layer.}
\label{fig:fig3}
\vspace{-0cm}
\end{figure*}

\subsection{Experimental Results}
\label{subsec:4.2}
To show the superiority of the proposed method, we conducted the preliminary experiments on the CIFAR-10 and CIFAR-100 datasets. In this paper, we trained the network three times from scratch to prove that the performance improvement is not due to the lucky weight initialization. Table~\ref{table1} presents the experimental results that compare the GAN performance between the proposed method and conventional ones including SNGAN~\cite{miyato2018spectral, miyato2018cgans}, BigGAN~\cite{brock2018large}, and GRB-GAN~\cite{park2021GRB}. In our experiments, we set the bottle-neck ratio in GBG-GAN as two. As summarized in Table~\ref{table1}, on both CIFAR-10 and CIFAR-100 datasets, the proposed method shows superior performance compared with the conventional methods that employ the standard residual block, \ie SNGAN and BigGAN, in terms of FID and IS. In addition, the proposed method outperforms the recent GAN technique, \ie GRB-GAN, which applies the newly designed residual block. These experimental results indicate that the proposed method is more effective than the conventional methods for the GAN-based image generation task. To ensure the ability of the proposed method, we also compared the performance of the proposed method with those of the conventional methods in the cGAN framework. As summarized in Table~\ref{table2}, the proposed method shows the outstanding performance compared to the conventional methods. In particular, the proposed method exhibits better performance than its counterparts by a large margin. For instance, on the CIFAR-100 dataset, the proposed method achieves the FID of 10.45, which is about 28.52\%, 20.41\%, and 16.33\% better than SNGAN, BigGAN, and GRB-GAN, respectively. Based on these observations, we believe that the proposed method can be widely used to accomplish notable performance in both GAN and cGAN schemes.

\begin{table*}[t]
\caption{GAN performance of the residual block with conventional gating shortcut methods on the CIFAR-10 and CIFAR-100 datasets in terms of IS and FID.}
\begin{center}
\begin{tabularx}{0.95\textwidth}{ |c| *{12}{Y|} }

\hline
\multicolumn{2}{|c|}{Shortcut name} & \multicolumn{2}{c|}{EGS} & \multicolumn{2}{c|}{SOG} & \multicolumn{2}{c|}{EGS with Conv.} & \multicolumn{2}{c|}{SOG with Conv.} & \multicolumn{2}{c|}{Proposed}\\
\hline

Dataset & & IS$\uparrow$ & FID$\downarrow$ & IS$\uparrow$ & FID$\downarrow$ & IS$\uparrow$ & FID$\downarrow$ & IS$\uparrow$ & FID$\downarrow$ & IS$\uparrow$ & FID$\downarrow$ \\
\hline
\multirow{5}*{CIFAR-10} & trial 1 & 7.73 & 13.59 & 7.77 & 14.04 & 7.89 & 12.45 & 7.82 & 12.74 & 7.93 & 10.94 \\
& trial 2 & 7.82 & 13.30 & 7.88 & 12.74 & 7.96 & 12.83 & 7.83 & 13.86 & 7.95 & 11.14 \\
& trial 3 & 7.89 & 13.03 & 7.83 & 13.32 & 7.94 & 12.52 & 7.76 & 13.75 & 8.00 & 10.88 \\
\cline{2-12}
& Mean & 7.81 & 13.31 & 7.83 & 13.37 & 7.93 & 12.60 & 7.80 & 13.45 &7.96 & 10.98 \\
\cline{2-12}
& Std & 0.08 & 0.28 & 0.06 & 0.65 & 0.03 & 0.20 & 0.04 & 0.62 & 0.04 & 0.14 \\
\hline
\multirow{5}*{CIFAR-100} & trial 1 & 7.98 & 17.48 & 8.12 & 17.00 & 8.09 & 16.36 & 8.02 & 16.73 & 8.22 & 15.17 \\
& trial 2 & 8.12 & 16.86 & 8.09 & 17.19 & 8.13 & 16.37 & 8.04 & 16.53 & 8.31 & 15.00 \\
& trial 3 & 8.01 & 16.80 & 8.04 & 16.46 & 8.10 & 16.28 & 7.99 & 17.05 & 8.33 & 15.39 \\
\cline{2-12}
& Mean & 8.03 & 17.05 & 8.08 & 16.89 & 8.11 & 16.34 & 8.01 & 16.77 & 8.29 & 15.19 \\
\cline{2-12}
& Std & 0.07 & 0.38 & 0.04 & 0.38 & 0.02 & 0.05 & 0.02 & 0.26 & 0.06 & 0.19 \\

\hline
\end{tabularx}
\end{center}
\label{table4}
\end{table*}

Indeed, the proposed method needs additional network parameters due to the convolution layers in the gated shortcut. Therefore, one might perceive that the performance improvement is caused by the increased number of network parameters. To mitigate this concern, we conducted ablation studies that match the network parameters of the conventional methods with that of the proposed one. In our experiments, we increased the channel of the residual block of SNGAN and BigGAN, whereas decreased the channel number of GRB-GAN. As presented in Tables~\ref{table3}, the SNGAN and BigGAN exhibit slight better performance when increasing the channel of the residual block, whereas the GRB-GAN shows similar or weak performance when decreasing the channel of the residual block. Even the performances of SNGAN and BigGAN are improved, but still show poor performance compared with proposed method. In other words, this variation, \ie increasing the number of network parameters, does not contribute to any noticeable performance improvement. Thus, we believe that the performance improvement is caused by the gated shortcut, not by the additional network parameters.

\begin{table}[t]
\caption{Comparison of the GAN performance between the proposed method with conventional ones on the LSUN-church and -tower datasets in terms of IS and FID.}
\begin{center}
\begin{tabular}{| c | c | c | c | c |}
\hline
\multicolumn{2}{|c|}{Network} & SNGAN & BigGAN & Proposed \\
\hline
Dataset & & FID$\downarrow$ & FID$\downarrow$ & FID$\downarrow$ \\
\hline
 & trial 1 & 7.83 & 8.18 & 6.61 \\
 & trial 2 & 8.16 & 7.88 & 6.62 \\
LSUN- & trial 3 & 8.22 &  8.11 & 6.71 \\
\cline{2-5}
church & Mean & 8.07 & 8.06 & 6.65 \\
\cline{2-5} 
 & Std & 0.21 & 0.15 & 0.06 \\
\hline
 & trial 1 & 12.42 & 13.89 & 10.79 \\
 & trial 2 & 12.29 & 12.38 & 10.38 \\
LSUN- & trial 3 & 12.54 & 11.98 & 9.96 \\
\cline{2-5}
tower & Mean & 12.42 & 12.78 & 10.38 \\
\cline{2-5} 
 & Std & 0.12 & 0.99 & 0.42 \\
\hline
\end{tabular}
\end{center}
\label{table5}
\end{table}

\begin{figure}
\centering
\includegraphics[width=\linewidth]{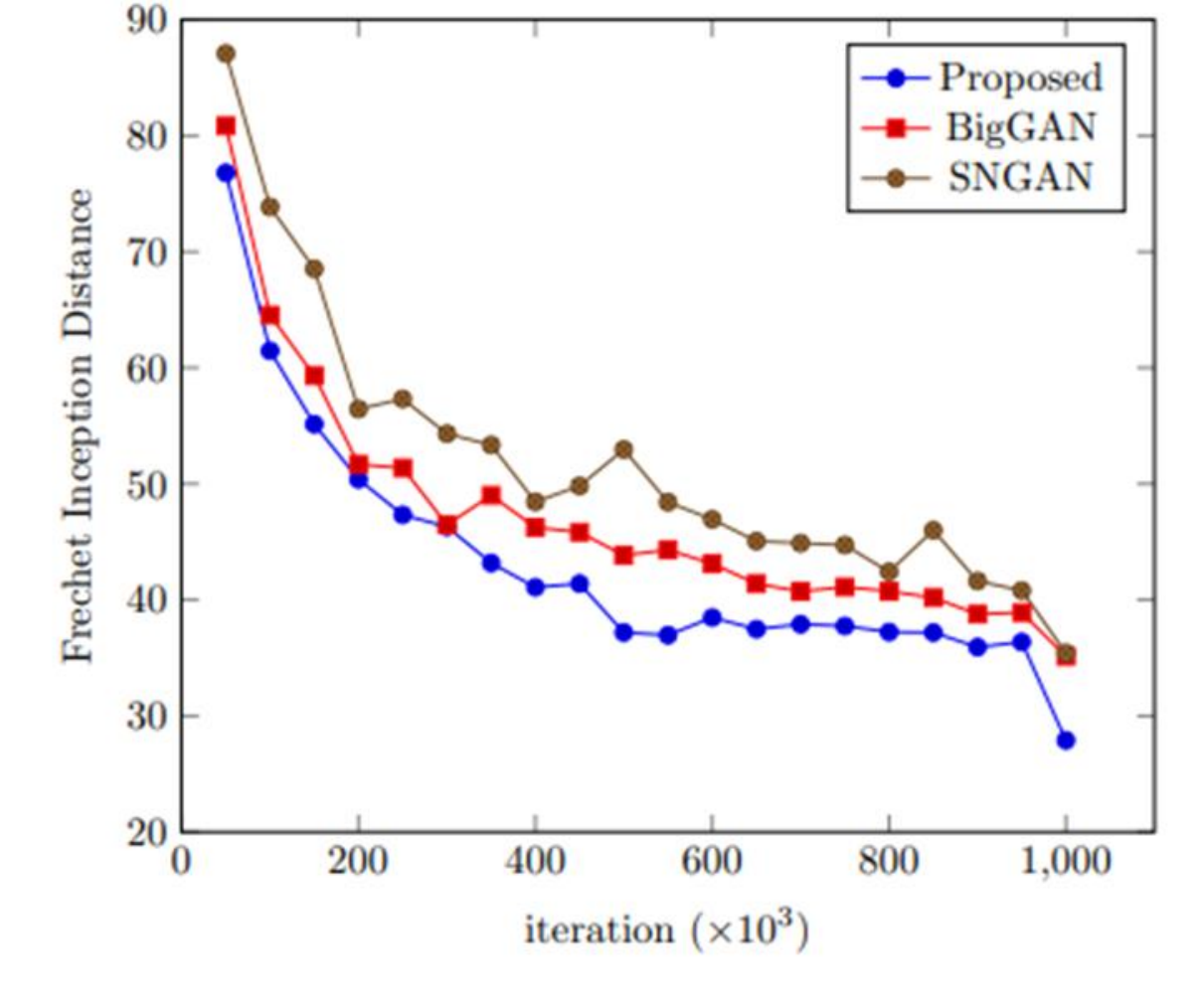}
\caption{Comparison of FID scores over training iterations. Blue, red, and yellow lines indicate the FID scores of proposed method, BigGAN, and SNGAN, respectively.}
\label{fig:fig_fid}
\vspace{-0cm}
\end{figure}

In addition, to demonstrate the effectiveness of the proposed gating mechanism, we compared the GAN performance of the proposed method with that of the previous gating shortcut methods in~\cite{he2016identity}. The previous gating shortcut methods are presented in Fig.~\ref{fig:fig3}, where Figs.~\ref{fig:fig3}(a) and (b) indicate the residual block with EGS and that with SOG, respectively. For a fair comparison with the proposed method, we added the previous gating shortcut techniques to the BigGAN residual block which infers the affine transformation parameters of BN from the noise vector. As described in Table~\ref{table4}, the conventional gating shortcut methods, \ie EGS and SOG, show better performance than the identity shortcut. Therefore, the identity shortcut may not the best choice for the generator; these results make our hypothesis reliable. Although EGS and SOG show fine performance compared to the identity shortcut, they still exhibit poor performance than the proposed method. That means, as mentioned in Section~\ref{subsec3.1}, these techniques are weight summation of the input and output features and do not fully carry out the advantage of the gating mechanism. Indeed, the proposed method contains a $1\times1$ convolution layer, \ie $\textrm{W}_o$, which produces $f_o$. Thus, one may suspect that the performances of EXG and SOG could be improved when there is $1\times1$ a convolutional layer at the end of the residual block. To resolve this concern, as depicted in Figs.~\ref{fig:fig3}(c) and (d), we built the EXG and SOG with an additional convolution layer and compared their performances with that of the proposed method. As shown in Table~\ref{table4}, even using the additional convolution layer, the proposed method still outperforms the modified conventional methods (\ie EGS with Conv. and SOG with Conv. in Table~\ref{table4}). Based on these results, we believe that the proposed method is further effective than the previous gating shortcut techniques.

\begin{figure}
\centering
\includegraphics[width=\linewidth]{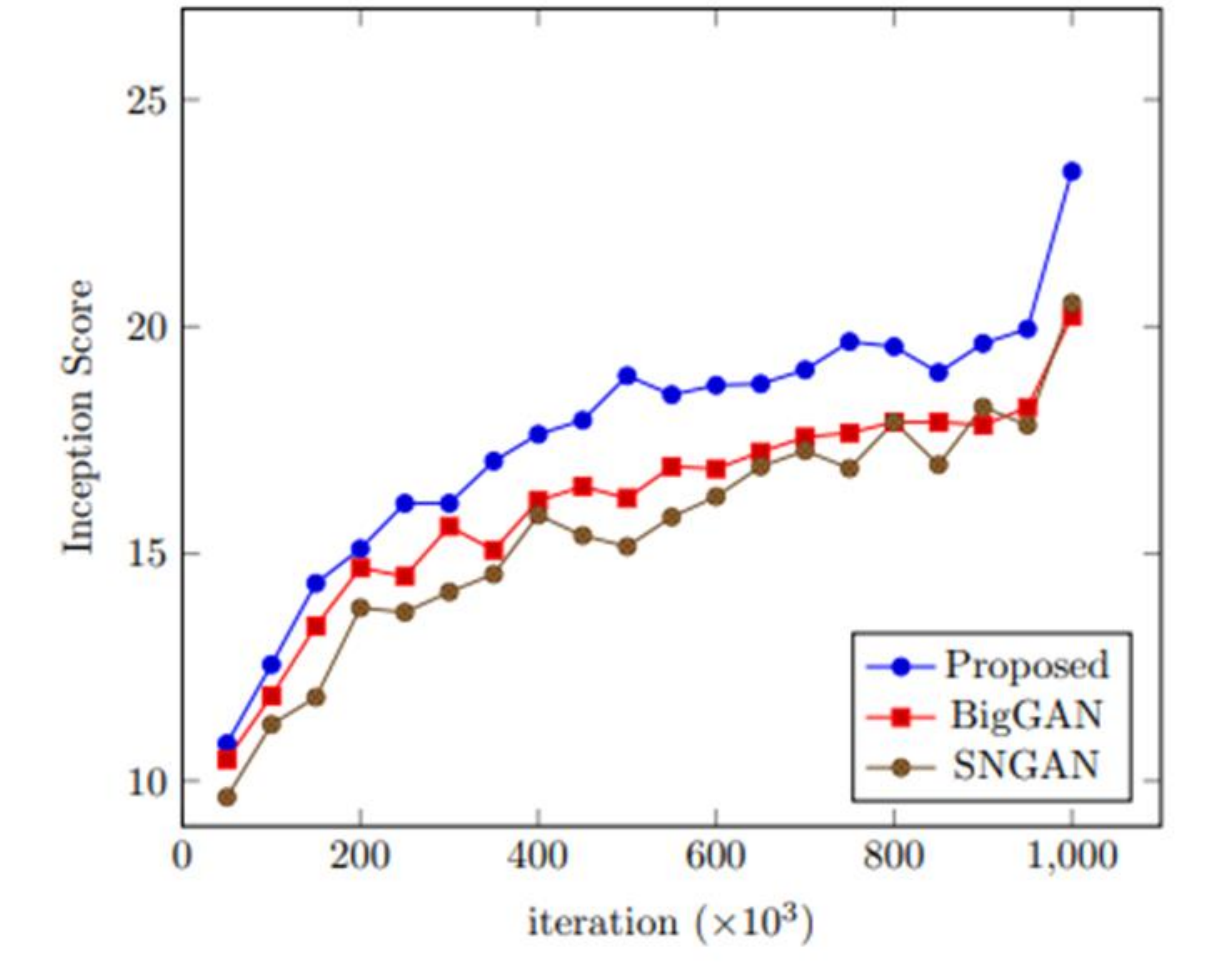}
\caption{Comparison of IS scores over training iterations. Blue, red, and yellow lines indicate the IS scores of proposed method, BigGAN, and SNGAN, respectively.}
\label{fig:fig_is}
\vspace{-0cm}
\end{figure}

\begin{table}[t]
\caption{Comparison of the final FID and IS scores on the tiny-ImageNet dataset.}
\begin{center}
\begin{tabular}{| c | c | c | c | c |}
\hline
Dataset & Metric & SNGAN & BigGAN & Proposed \\
\hline
tiny- & IS$\uparrow$ & 20.52 & 20.23 & 23.42\\
\cline{2-5}
ImageNet & FID$\downarrow$ & 35.42 & 35.13 & 27.90 \\

\hline
\end{tabular}
\end{center}
\label{table:table6}
\vspace{-0cm}
\end{table}

\begin{figure}
\centering
\includegraphics[width=0.95\linewidth]{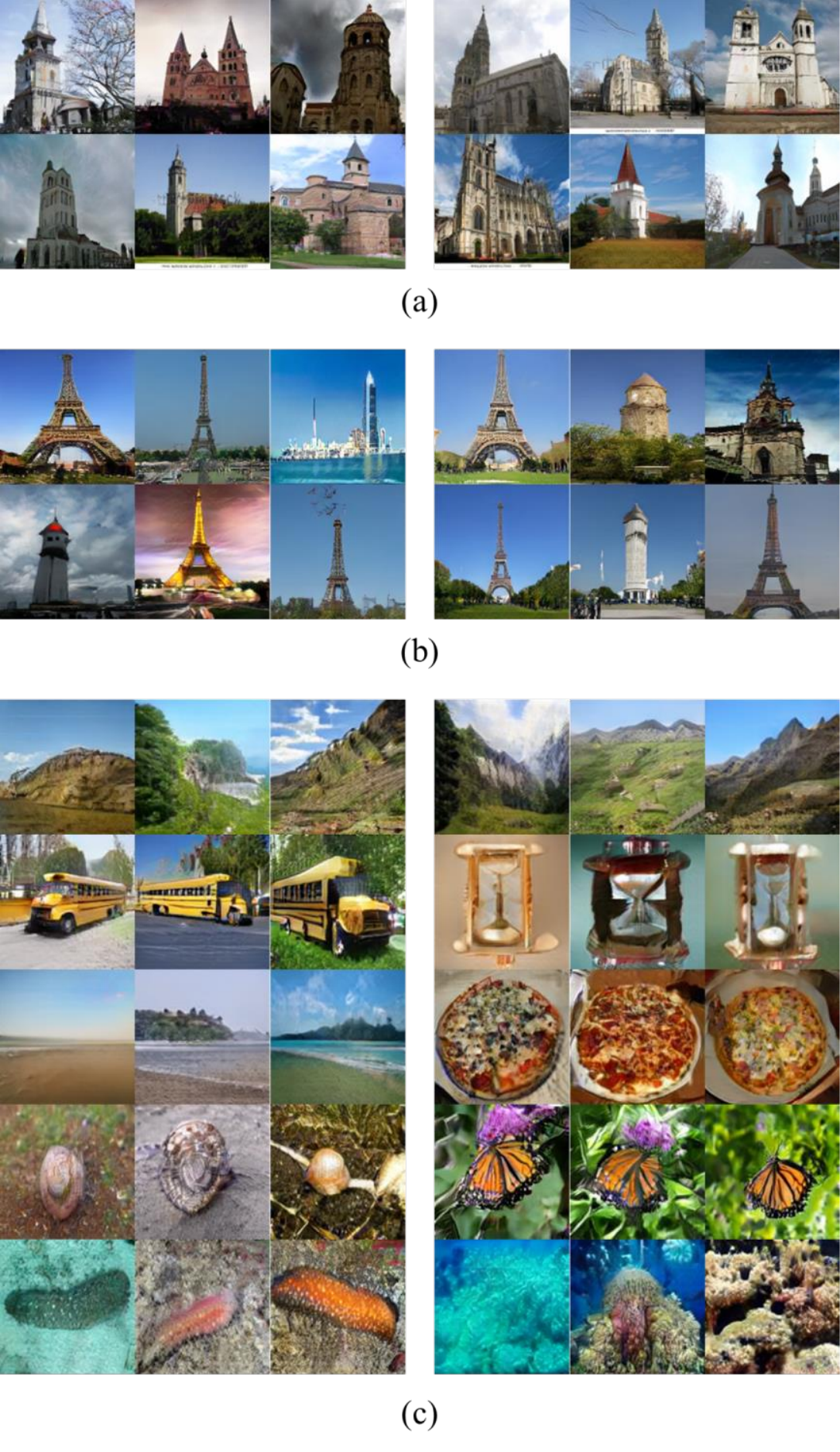}
\caption{Samples of the generated images. (a) Generated images on the LSUN-church dataset, (b) Generated images on the LSUN-tower dataset, (c) Generated images on the tiny-ImageNet dataset.}
\label{fig_gen}
\vspace{0cm}
\end{figure}

Furthermore, to prove the superiority of the proposed method when synthesizing images on challenging datasets, we trained the network on the LSUN-church and -tower datasets. More specifically, we trained the network following the experimental setting for GAN framework and applied the TTUR technique. It is worth noting that we attempted to train GRB-GAN with the TTUR technique, but it fails to train; the GRB-GAN only can be trained when updating the discriminator five times while the generator is updated once. Thus, we measured the performance of the proposed method against that of the conventional ones except for GRB-GAN. As presented in Table~\ref{table5}, on the challenging dataset, the proposed method exhibits outstanding performance compared to the conventional methods. Furthermore, we trained the network using the tiny-ImageNet dataset, following the experimental setting for the cGAN scheme. Since we already revealed that the performance improvement is not caused by the lucky weight initialization, we trained the network a single time from scratch. Instead, we drew graphs that present the FID and IS scores over training iterations and summarized the final scores in ~\ref{table:table6}. As depicted in Figs.~\ref{fig:fig_fid} and~\ref{fig:fig_is}, the proposed method shows steadily better performance than the conventional methods, \ie SNGAN and BigGAN, over the training procedure. In addition, as shown in Table~\ref{table:table6}, the proposed method exhibits the remarkable performance. Fig.~\ref{fig_gen} shows the samples of the generated images. As depicted in Fig.~\ref{fig_gen}, the proposed method is effective to synthesize the images with complex scenes. Based on these results, we concluded that the proposed method can generate visually pleasing images on the challenging datasets.

\section{Conclusion}
\label{sec5}
This paper has introduced a novel shortcut technique, called gated shortcut, which shows outstanding performance for GAN-based image generation task. By simply replacing the identity shortcut with the proposed method, the generator can synthesize high-quality image, results in boosting the GAN performance. The major strength point of the proposed method is that it can be easily applied to the existing residual block with marginal overheads, while achieving much lower FID and higer IS scores than the conventional methods. To prove this advantage, this paper deeply investigates the proposed method in various aspects through many-sided ablation studies. Therefore, we believe that the proposed is applicable to various GAN-based image generation tasks. Indeed, we agree that the proposed method works well for the GAN-based image generation task, but there might be some concerns in other GAN-based applications. Even though we only cover the effectiveness of the gated shortcut in the field of GAN-based image generation tasks, this paper has shown that the proposed method could achieve notable performance. As our future work, we plan to further investigate the effectiveness of the gated shortcut in other various applications.

\bibliographystyle{spmpsci}      
\bibliography{egbib.bib}

\vspace{100mm}

\begin{wrapfigure}{l}{0\textwidth}
\includegraphics[width=1in, height=1.25in]{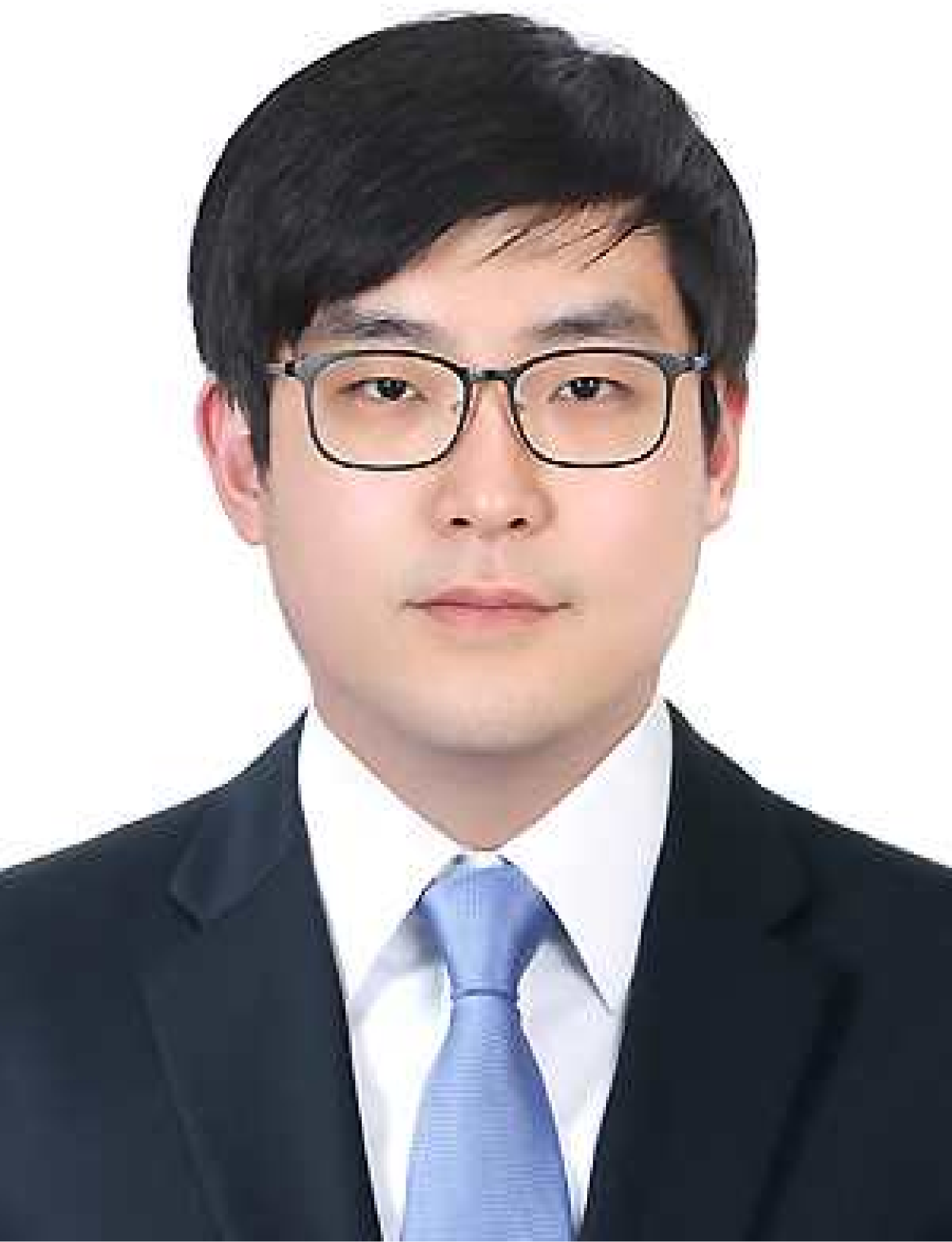}
\end{wrapfigure} 
\vspace{5mm}
\textbf{Seung Park} received the B.S. and Ph.D. degrees in electrical engineering from Korea University, Seoul, South Korea, in 2013 and 2020, respectively. He is currently a Clinical Assistant Professor in Chungbuk National University Hospital. His current research interests include computer vision and image processing.

\begin{wrapfigure}{l}{0\textwidth}
\includegraphics[width=1in, height=1.25in]{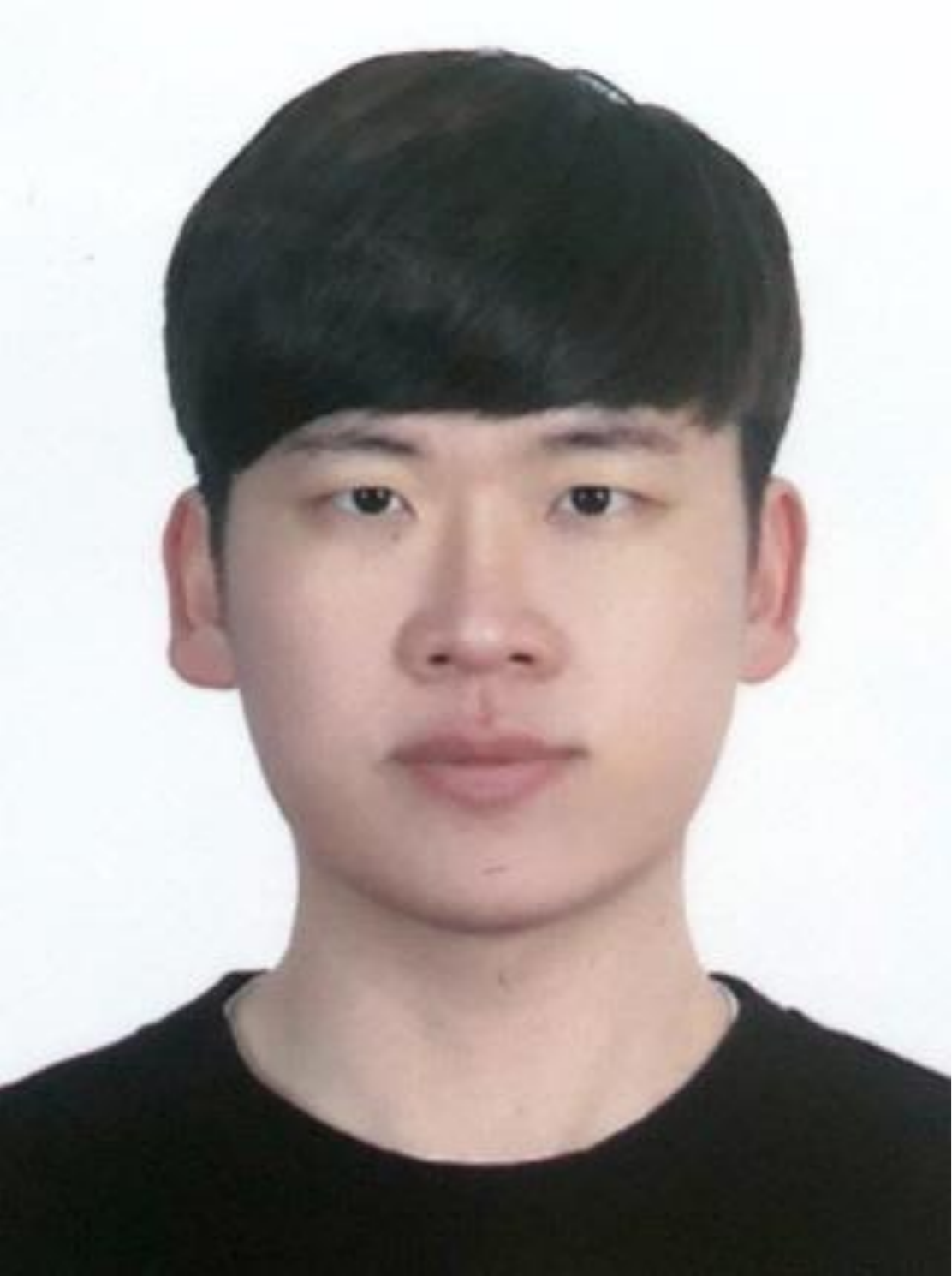}
\end{wrapfigure} 
\vspace{5mm}
\textbf{Cheol-Hwan Yoo} received the B.S. and Ph.D. degrees in electrical engineering from Korea University, Seoul, South Korea, in 2014 and 2020, respectively. Since 2020, he has been with the Electronics and Telecommunications Research Institute (ETRI), South Korea, as a Researcher. His research interests include deep learning, image processing, computer vision, and human-robot interaction.

\begin{wrapfigure}{l}{0\textwidth}
\includegraphics[width=1in, height=1.25in]{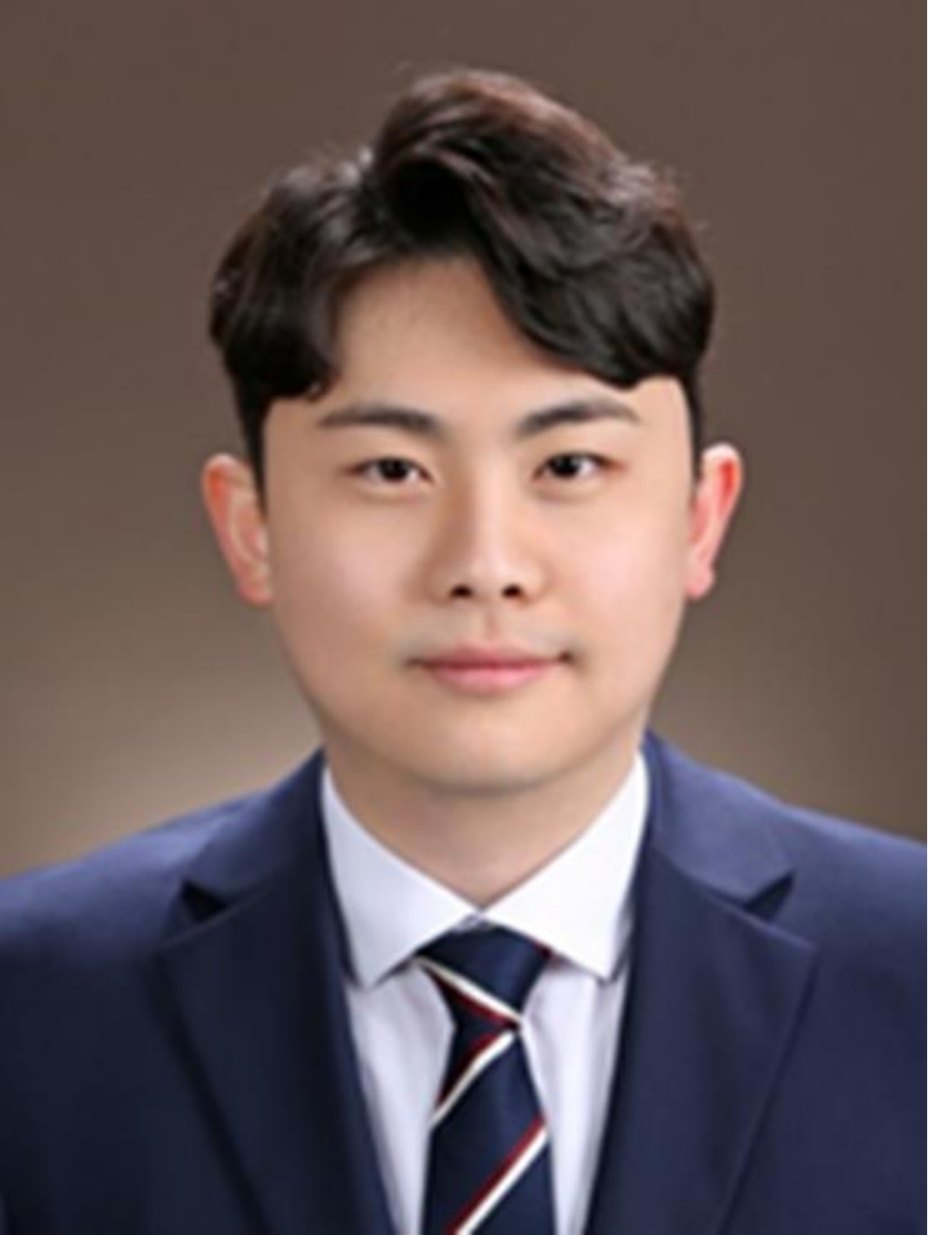}
\end{wrapfigure} 
\vspace{5mm}
\textbf{Yong-Goo Shin} received the B.S. and Ph.D. degrees in electronics engineering from Korea University, Seoul, South Korea, in 2014 and 2020, respectively. He is currently an Assistant Professor with the Division of Smart Interdisciplinary Engineering, Hannam University. He has published over 10 international journal articles in fields such as image processing, computer vision, and deep learning.



\end{document}